\documentclass{article}

\usepackage{PRIMEarxiv}

\usepackage[utf8]{inputenc} 
\usepackage[T1]{fontenc}    
\usepackage{hyperref}       
\usepackage{url}            
\usepackage{booktabs}       
\usepackage{amsfonts}       
\usepackage{nicefrac}       
\usepackage{microtype}      
\usepackage{lipsum}
\usepackage{fancyhdr}       
\usepackage{graphicx}       
\graphicspath{{media/}}     

\usepackage{amsmath}
\usepackage{amssymb}
\usepackage{multirow}
\usepackage{algorithm}
\usepackage{algorithmic}
\usepackage{color}

\title{Synthesis-based Imaging-Differentiation Representation Learning for Multi-Sequence 3D/4D MRI
}

\author{
  Luyi Han, Tao Tan\thanks{\textit{Corresponding Author}: Tao Tan (taotanjs@gmail.com)}, Tianyu Zhang, Xin Wang, Yuan Gao, Jonas Teuwen, Ritse Mann \\
  Department of Radiology \\
  The Netherlands Cancer Institute (NKI) \\
  Amsterdam \\
  \texttt{\{l.han, t.tan, t.zhang, x.wang, y.gao, j.teuwen, r.mann\}@nki.nl} \\
   \And
  Yunzhi Huang \\
  School of Automation \\
  Nanjing University of Information Science and Technology \\
  Nanjing \\
  \texttt{yunzhi.huang.scu@gmail.com} \\
}

\begin{document}
\maketitle

\begin{abstract}
Multi-sequence MRIs can be necessary for reliable diagnosis in clinical practice due to the complimentary information within sequences. 
However, redundant information exists across sequences, which interferes with mining efficient representations by modern machine learning or deep learning models.
To handle various clinical scenarios, we propose a sequence-to-sequence generation framework (Seq2Seq) for imaging-differentiation representation learning.
In this study, not only do we propose arbitrary 3D/4D sequence generation within one model to generate any specified target sequence, but also we are able to rank the importance of each sequence based on a new metric estimating the difficulty of a sequence being generated.
Furthermore, we also exploit the generation inability of the model to extract regions that contain unique information for each sequence.
We conduct extensive experiments using three datasets including a toy dataset of 20,000 simulated subjects, a brain MRI dataset of 1,251 subjects, and a breast MRI dataset of 2,101 subjects, to demonstrate that (1) our proposed Seq2Seq is efficient and lightweight for complex clinical datasets and can achieve excellent image quality; (2) top-ranking sequences can be used to replace complete sequences with non-inferior performance; (3) combining MRI with our imaging-differentiation map leads to better performance in clinical tasks such as glioblastoma MGMT promoter methylation status prediction and breast cancer pathological complete response status prediction.
Our code is available at \url{https://github.com/fiy2W/mri_seq2seq}.
\end{abstract}


\section{Introduction}
\label{sec:introduction}
Multi-sequence MRIs show different characteristics of protons within tissues, resulting in various image appearances of water and fat tissue when using particular settings of radiofrequency pulses and gradients~\footnote{\url{https://doi.org/10.53347/rID-37346}}. Clinicians generally utilize multi-sequence MRIs to reach a reliable diagnosis because a single sequence is often insufficient to describe lesions depending on the clinical tasks. For instance, the standard clinical scan protocol for glioblastoma includes T1-weighted (T1), contrast-enhanced (T1Gd), T2-weighted (T2), and T2-fluid-attenuated inversion recovery (Flair) sequences~\cite{shukla2017advanced}. Dynamic contrast-enhanced (DCE) and diffusion-weighted imaging (DWI) are required in the current standard protocol for breast MRI imaging, which is, for example, used to predict the response to neoadjuvant chemotherapy (NAC)~\cite{chen2013clinical}. Because of their unique cellular microenvironment, lesions may be recognized and classified based on the different appearance combinations in multi-sequence MRI. Diagnosis of diseases from these combinations, therefore, relies on the summarization of multiple features, which requires extensive clinical experience. The diagnostic reliability of multi-sequence MRIs may still be improved, as it is often unclear what the most important information is to use for lesion characterization.

\begin{figure}[t]
    \centering
    \includegraphics[width=0.5\linewidth]{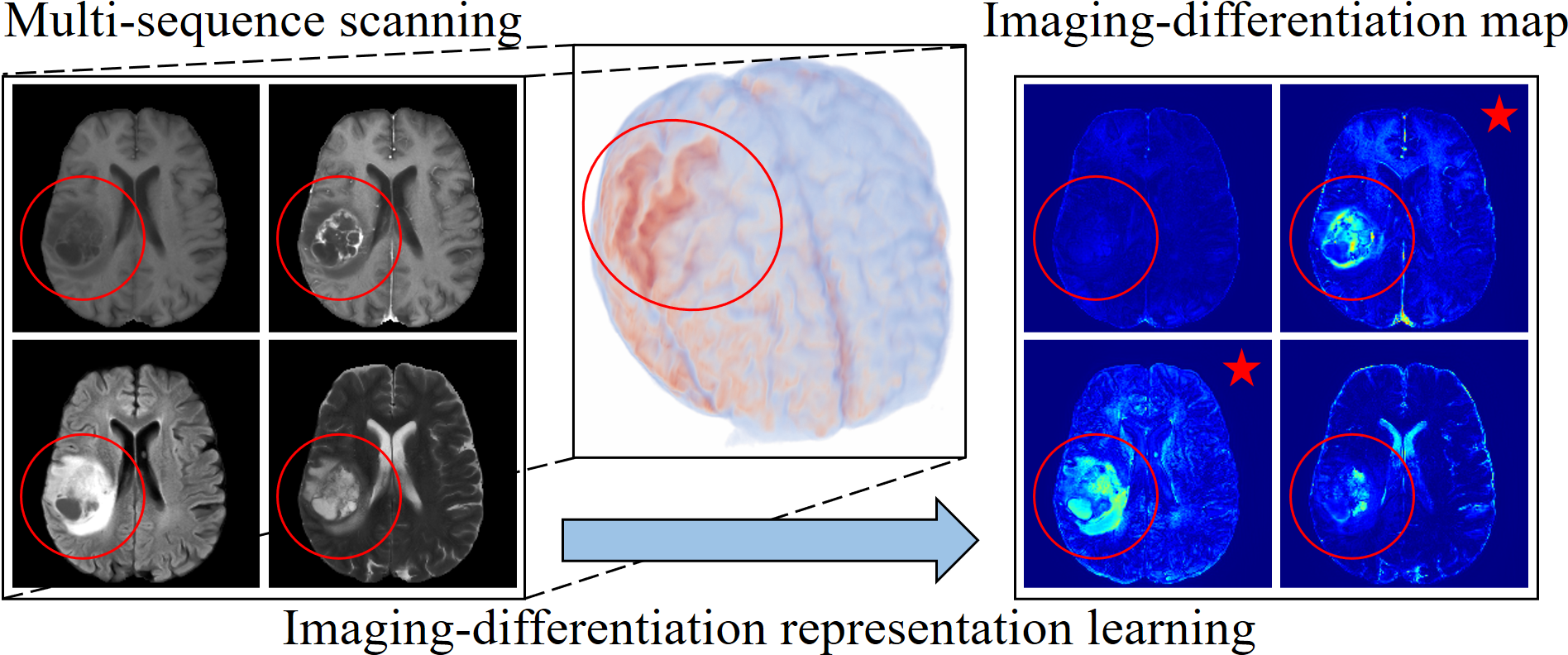}
    \caption{In clinical, multi-sequence scanning is utilized for a reliable diagnosis. Our proposed imaging-differentiation representation learning approach can rank the importance of sequences and highlight specific imaging-differentiation regions. Red circles contour the glioblastoma region, and red $\star$ marks the top 2 important sequences.}\label{fig:fig1}
\end{figure}

In recent years, convolutional neural networks (CNN) have been widely used for processing multi-sequence MRI~\cite{feng2020knowledge,grovik2020deep,tang2020postoperative,zhuang2022cardiac}.
Different from a single sequence pattern analysis, however, multi-sequence CNN-based research has the following challenges: (1) limited samples but with redundant information across inter sequences; and (2) CNNs trained using all sequences cannot handle missing sequences in clinical datasets.
In practice, the MRI sequences are often incomplete due to high scanning costs and also time constraint and different acquisition protocols between hospitals.
Missing sequence synthesis is a solution to handle these issues~\cite{li2019diamondgan,sharma2019missing,zhou2020hi,jung2021conditional,uzunova2020memory,dalmaz2022resvit,li2022virtual}.
By leveraging synthesis, Han \textit{et al.}~\cite{han2018gan} generate realistic multi-sequence brain MRIs for data augmentation purposes.
Campello \textit{et al.}~\cite{campello2019combining} utilize synthetic sequences to improve the performance of cardiac segmentation.
However, sequences containing highly specific information, \textit{e.g.} T1Gd, are expected hard to be synthesized~\cite{sharma2019missing}.
Generated images are not reliable and difficult to convincingly be applied for clinical diagnosis.

In practice, radiologists read images following guidelines for a specific disease, evaluating different manifestations from different sequences based on clinical experience. Rather than blindly generating and using synthetic MRI sequences, we introduce an imaging-differentiation representation learning approach for multi-sequence MRI learning as shown in Fig.~\ref{fig:fig1}.
Our contributions are three folds:
\begin{itemize}
    \item[i] We propose a simple and efficient end-to-end model to transform an arbitrary given 3D/4D MRI sequence into a target sequence.
    \item[ii] We rank the importance of each sequence contributing to the MRI synthesis for non-inferiority clinical scanning selection based on our proposed new metric.
    \item[iii] We quantitatively estimate the amount of incremental information of each sequence compared to the remaining sequences, and further employ it to guide the multi-sequence learning in specific clinical tasks.
\end{itemize}

\section{Related Work}
\label{sec:relatedwork}
\subsection{Reliability Learning}
Recently, some researchers consider the reliability and contribution of each sequence in their works.
Qi \textit{et al.}~\cite{qi2020multi} propose a conditional GAN to synthesize CT based on multi-sequence MRI. They compare combinations of input sequences to verify their contributions to the synthesis. 
By ablation study of the combinations, the contribution of each sequence can be evaluated indirectly but cannot be analyzed quantitatively.
Huang \textit{et al.}~\cite{huang2022evidence} utilize the formalism of Dempster-Shafer theory to fuse segmentation results from multi-sequence MRI, providing the reliability of different sequences to different segmentation labels. Their findings are highly consistent with clinical experience.
However, this method is label-driven, which needs additional annotations of segmentation. It also cannot highlight specific contributing areas from each sequence.
Finck \textit{et al.}~\cite{finck2022uncertainty} translate T1 and Flair to high-contrast double inversion recovery (DIR) images through GAN and outputting with an uncertainty map by utilizing dropout in the model. The proposed uncertainty maps can show some lesion-specific regions and enhance their clinical utility and trustworthiness. But they focus more on regions with high-frequency information, which are easy to be influenced by random dropouts.
This motivates us to propose a simple and efficient approach to analyze sequence contribution in image synthesis and further highlight specific contributing regions.
As image-to-image translation should not create nonexistent information, we take advantage of this to determine imaging-differentiation regions for each sequence.

\subsection{Hyperparameter Network}
To handle clinical settings with missing sequences, we merge the arbitrary sequence generation into a single model, which can learn more correlation information from multiple sequences.
The hyperparameter network is a model whose output can be dynamically conditioned by a control signal. In the style transfer task, it is convenient and efficient to use hyperparameter networks to transfer source styles to multiple target styles. Specifically, hyperparameter networks can be divided into two groups: (1) adaptive normalization~\cite{huang2017arbitrary} and (2) dynamic filters~\cite{shen2018neural}.
Huang \textit{et al.}~\cite{huang2017arbitrary} first propose the adaptive instance normalization (AdaIN) layer to match the mean and variance of the content features between the source style and the target style. This idea has been applied to many style transfer works~\cite{huang2018multimodal,lee2018diverse,choi2018stargan}. AdaIN is a lightweight module but sometimes can cause water droplet-like artifacts as the generator is capable of controlling the statistics of AdaIN easily by creating strong and localized spikes~\cite{karras2020analyzing}.
In contrast, a dynamic filter can resolve these artifacts. Jia \textit{et al.}~\cite{jia2016dynamic} first propose dynamic filter networks, whose filters are generated from inputs.
Shen \textit{et al.}~\cite{shen2018neural} propose a meta-network to build up the transformation network by giving a target style image.
Similarly, Liu \textit{et al.}~\cite{liu2019learning} utilize the semantic segmentation mask to predict the weights of convolutional kernels in the generator, achieving better semantic layouts in the synthetic image.
However, dynamic filter-based methods require a large number of weights in the kernel branch, which is usually computationally expensive and prone to overfitting in limited datasets~\cite{zhao2018dynamic,wu2018dynamic}.

\section{Proposed Approach}
\label{sec:method}
\subsection{Overview}
Multi-sequence MRI provides both redundant (delineated in most sequences) and imaging-differentiation (demarcated in seldom sequences) information for the tissues. Leveraging redundant information between sequences is the key to MRI synthesis methods~\cite{sharma2019missing}.
Based on this, we propose a sequence-to-sequence (Seq2Seq) generator to realize a rapid transformation between two arbitrary sequences and extract the redundant information from each sequence. To also handle 4D protocols of MRI, we employ long short-term memory (LSTM) mechanism to capture longitudinal information from 4D MRIs. In addition, to generate various sequences in a single model, hyper-parameter blocks related to the target sequence code are used in the network architecture.
To separate imaging-differentiation from redundant information, accumulated errors between the real target sequence and synthetic sequences generated from the remaining sequences describe the corresponding imaging-differentiation map, which is valuable for clinical applications.

\begin{figure}[!htbp]
	\centering
	\includegraphics[width=0.95\linewidth]{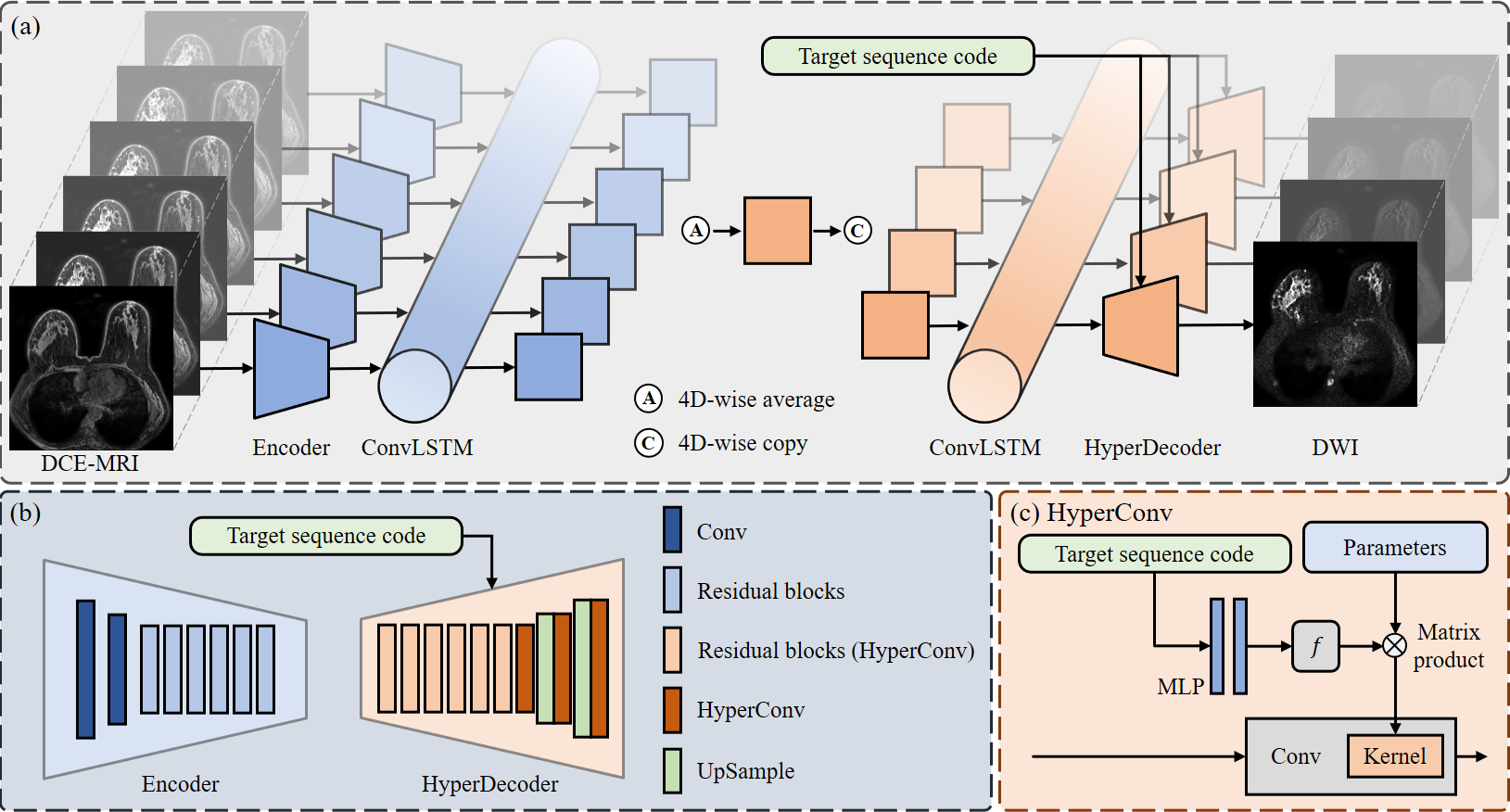}
	\caption{The overview of the proposed framework. (a) The architecture of Seq2Seq generator inputting with 3D/4D MRI sequence and outputting with target sequence; (b) the structure of Encoder and HyperDecoder; and (c) the structure of HyperConv layer.} \label{fig:framework}
\end{figure}

\subsection{Seq2Seq generator}
To ensure that the model can adapt to both 3D and 4D MRIs, we compress different input sequences into an identical representation, and then reconstruct arbitrary target sequences based on the identical representation. 
As illustrated in Fig.~\ref{fig:framework} (a), the proposed Seq2Seq generator involves an encoding path $\mathbf{E}$ and a conditional decoding path $\mathbf{G}$.
In the encoding path, an Encoder is first utilized to extract features from 3D MRI images. To handle 4D MRIs with multiple timepoints/parameters, we introduce multiple Encoders with shared weights for 4D MRIs and then capture spatiotemporal correlations between the images by ConvLSTM~\cite{shi2015convolutional} layer. To further compress the representation, we fuse output features from different timepoints or parameters by 4D-wise average.
In the conditional decoding path, the compressed representation is first expanded to the same number of timepoints or parameters as the target sequence by the 4D-wise copy operation. Then, the ConvLSTM~\cite{shi2015convolutional} layer can disentangle the spatiotemporal correlations in the compressed representation and recover images for each timepoint or parameter. Finally, the HyperDecoder which receives both features and sequence code is utilized to reconstruct the target sequence. Similarly, multiple share-weighted HyperDecoders are employed to deal with 4D MRI output.

Fig.~\ref{fig:framework} (b) shows the details of the Encoder and HyperDecoder. The Encoder is simply implemented consisting of two convolutional layers and six residual blocks. Two convolutional layers can 4 times downsample images, and residual blocks can extract the high-level representation.
HyperDecoder involves six Hyper-residual blocks and three HyperConv layers.
The HyperConv layer is the key to generating target sequences with a given sequence code, details of which are in Section~\ref{sec:hyperconv}. Hyper-residual blocks are residual blocks by replacing the convolutional layer with the HyperConv layer. Note that, to reconstruct the image with the same size as input, there are two $2\times$ upsample layers before the last two HyperConv layers, respectively.

\subsection{HyperConv layer}
\label{sec:hyperconv}
The convolutional layer is the basic structure of the deep learning network.
We employ one-hot encoding to present different sequence codes and expect to utilize it to control the output of convolutional layers.
Inspired by Zhao~ \textit{et al.}~\cite{zhao2018dynamic}, we further optimize its implementation for a more lightweight module--HyperConv layer. As shown in Fig.~\ref{fig:framework} (c), we first predict a mapping function $f$ from the target sequence code through the multilayer perceptron (MLP). Then, function $f$ is capable of mapping the trainable weight bank to a specific convolutional kernel of the target sequence by matrix product.
In this way, the shared information of multi-sequence MRI can be saved in the parameter bank, and it is also easier for the layer to switch to kernels targeting different sequences.
The pseudocode implementation of the HyperConv layer is described in Algorithm~\ref{alg:hyperconv}.

\begin{algorithm}
    \caption{PyTorch-like pseudocode of HyperConv}
    \label{alg:hyperconv}
    \begin{algorithmic}
        \REQUIRE $x$: input feature maps; $s$: target sequence code;
        \STATE $c_s$: code dimension; $c_w$: weight bank dimension;
        \STATE kshape: kernel shape of the convolutional layer
        \ENSURE $y$: output feature maps
        \STATE param = \textbf{Parameter}(*kshape, $c_w$)
        \STATE $f$ = \textbf{Linear}($c_s$, $c_w$).forward($s$).view($c_w$, 1)
        \STATE kernel = \textbf{matmul}(param, $f$).view(*kshape)
        \STATE $y$ = \textbf{conv}($x$, kernel)
    \end{algorithmic}
\end{algorithm}

\subsection{Loss function}
We assume that all the patients theoretically acquire complete multi-sequence MRIs (\textit{e.g.} T1, T2, DWI, DCE), which are represented as $\mathcal{D}=\{X^{(n)}_1, X^{(n)}_2, ..., X^{(n)}_S\}_{n=1}^{N}$.
Note that, $\mathcal{D}$ is a set of $N$ subjects with $S$ sequences for each.
To mimic clinical setting with missing sequences, we define the indicator function $\mathbf{1}_{\mathcal{A}}: \mathcal{D}\rightarrow\{0,1\}$ to represent the incomplete subset $\mathcal{A}$ with only available sequences from $\mathcal{D}$.
With the given available sequences from $\mathcal{A}^{(n)}$, each available sequence $X_i$ is encoded in features by the share-weighted encoding path $\mathbf{E}$, and then the features are decoded to a target sequence by the conditional decoding path $\mathbf{G}$.
To restrict the reconstruction similarity between generated sequences and the real ones, a supervised reconstruction loss is given as,
\begin{equation}
    \label{eq:reconstruction}
    \begin{aligned}
        \mathcal{L}_{rec}=\sum_{X_i\in\mathcal{A}^{(n)}}\sum_{X_j\in\mathcal{A}^{(n)}}\lambda_{r}\cdot\|X'_{i\rightarrow j}-X_j\|_1 + \lambda_{p}\cdot\mathcal{L}_{p}(X'_{i\rightarrow j}, X_j)
    \end{aligned}
\end{equation}
where $X'_{i\rightarrow j}=\mathbf{G}(\mathbf{E}(X_i)|s_{j})$, $s_j$ indicates the sequence code of $X_j$, $\| \cdot \|_1$ is a $L_1$ loss, and $\mathcal{L}_{p}$ refers to the perceptual loss based on pre-trained VGG19. $\lambda_{r}$ and $\lambda_{p}$ are weight terms and are experimentally set to be $10$ and $0.01$.

However, for those with severe missing data, especially completely unpaired data, Eq.~\ref{eq:reconstruction} will degenerate into the sequence self-reconstruction form.
And the model is unable to learn the common representation between different sequences.
Thus, we employ the adversarial loss and the cycle-consistent loss~\cite{zhu2017unpaired} to force the image translation and capture the mutual representation.
\begin{equation}
    \label{eq:adversarial}
    \begin{aligned}
        \min_{\mathbf{D}}\max_{\mathbf{G}}\mathcal{L}_{adv}=\sum_{X_i\in\mathcal{A}^{(m)}}{\sum_{j\neq i}^{X_j\in\mathcal{A}^{(n)}}} \|\mathbf{D}_j(X_j)-1\|_2+\|\mathbf{D}_j(X'_{i\rightarrow j})\|_2
    \end{aligned}
\end{equation}
\begin{equation}
    \label{eq:cycle}
    \begin{aligned}
        \mathcal{L}_{cyc}=\sum_{X_i\in\mathcal{A}^{(n)}}{\sum_{j\neq i}^{S}{\|X''_{i\rightarrow j\rightarrow i}-X_i\|_1}}
    \end{aligned}
\end{equation}
where $X''_{i\rightarrow j\rightarrow i}=\mathbf{G}(\mathbf{E}(X'_{i\rightarrow j})|s_i)$, $\| \cdot \|_2$ is a $L_2$ loss, $\mathbf{D}_j$ indicates the discriminator for sequence $j$.

\subsection{Imaging-differentiation representation}
We employ the difficulty of one sequence being generated from other sequences to present its imaging differentiation and refer to the ability of one sequence to generate other different sequences as its totipotency.
Compared with common (easy to be generated) and low-totipotent sequences, the imaging-differentiation and high-totipotent sequences make an irreplaceable contribution to the multi-sequence analysis.
Here, we define the degree centrality of totipotency $\mathcal{C}_{t}$ and imaging differentiation $\mathcal{C}_{d}$ for each type of sequence, respectively,
\begin{equation}
    \label{eq:degree-centrality}
    \begin{aligned}
        \mathcal{C}_{t}(i)&=\frac{1}{S}\sum_{j=0}^{S}{A_{ij}} \\
        \mathcal{C}_{d}(i)&=-\frac{1}{S}\sum_{j=0}^{S}{A_{ji}} \\
    \end{aligned}
\end{equation}
where $A_{ij}$ is the adjacency matrix to evaluate the similarity between generated image and ground truth, here we set $A_{ij}={\rm nPSNR}_{ij}+{\rm nSSIM}_{ij}-{\rm nLPIPS}_{ij}$. ${\rm nPSNR}$, {\rm nSSIM}, and {\rm nLPIPS} refer to the normalized metrics. $i$ and $j$ indicate the source and target sequence, respectively.

Besides sequence-level analysis, we can also achieve pixel-level representation.
The areas of one sequence, that are difficult to be generated by other sequences, are considered to be the imaging-differentiation regions. Based on this, we define the imaging-differentiation map of a given sequence as follows,
\begin{equation}
    \label{eq:imaging-differentiation-map}
    \begin{aligned}
        \mathcal{M}_{d}(X_i)=\frac{1}{S}\sum_{j=0}^{S}{\left|X'_{j\rightarrow i}-X_i\right|} \\
    \end{aligned}
\end{equation}

The imaging-differentiation map can highlight the most worthy attention region of a sequence, which cannot be deduced from other sequences. This can guide the model to focus on more representative regions rather than those with redundant information.
For multi-sequence analysis, the imaging-differentiation map is capable of improving the performance of the model and reducing distractions.

\section{Experimental Results}
\label{sec:result}
\subsection{Datasets and setting}
\begin{figure}[b]
    \begin{minipage}{0.1\linewidth}
        \centerline{Case 1}
    \end{minipage}
    \begin{minipage}{0.21\linewidth}
        \centerline{\includegraphics[width=\textwidth]{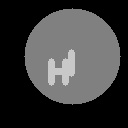}}
    \end{minipage}
    \begin{minipage}{0.21\linewidth}
        \centerline{\includegraphics[width=\textwidth]{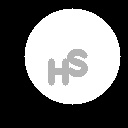}}
    \end{minipage}
    \begin{minipage}{0.21\linewidth}
        \centerline{\includegraphics[width=\textwidth]{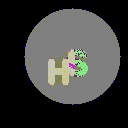}}
    \end{minipage}
    \begin{minipage}{0.21\linewidth}
        \centerline{\includegraphics[width=\textwidth]{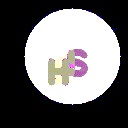}}
    \end{minipage}
    
    \vspace{2pt}
    
    \begin{minipage}{0.1\linewidth}
        \centerline{Case 2}
    \end{minipage}
    \begin{minipage}{0.21\linewidth}
        \centerline{\includegraphics[width=\textwidth]{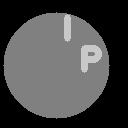}}
    \end{minipage}
    \begin{minipage}{0.21\linewidth}
        \centerline{\includegraphics[width=\textwidth]{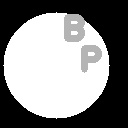}}
    \end{minipage}
    \begin{minipage}{0.21\linewidth}
        \centerline{\includegraphics[width=\textwidth]{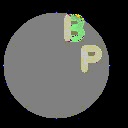}}
    \end{minipage}
    \begin{minipage}{0.21\linewidth}
        \centerline{\includegraphics[width=\textwidth]{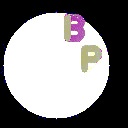}}
    \end{minipage}
    
    \vspace{2pt}
    
    \begin{minipage}{0.1\linewidth}
        \centerline{}
    \end{minipage}
    \begin{minipage}{0.21\linewidth}
        \centerline{$X_1$}
    \end{minipage}
    \begin{minipage}{0.21\linewidth}
        \centerline{$X_2$}
    \end{minipage}
    \begin{minipage}{0.21\linewidth}
        \centerline{$\mathcal{M}_{d}(X_1)$}
    \end{minipage}
    \begin{minipage}{0.21\linewidth}
        \centerline{$\mathcal{M}_{d}(X_2)$}
    \end{minipage}
	
	\caption{Examples of the toy dataset. The labels of $X_1$ and $X_2$ are ``I" and ``S" in case 1, and ``I" and ``B" in case 2. $\mathcal{M}_{d}$ indicates the imaging-differentiation map generated by our proposed method. For better visualization, we overlap the original image with the heatmap of $\mathcal{M}_{d}$.} \label{fig:prototype}
\end{figure}

\paragraph{Toy dataset}
We simulate a toy dataset to explain the role of imaging-differentiation representation learning. As shown in Fig.~\ref{fig:prototype}, there are two sequences with a size of $128\times128$ in each subject. Both sequences consist of a circle and two alphabets in identical locations but in different colors, indicating the same tissue with different intensities in different MRI sequences. To present common and imaging-differentiation tissue, respectively, one of the paired alphabets is the same, and another pair is different. And the target of this dataset is to recognize the different alphabets between two sequences. We randomly select 9,000 subjects for training, 1,000 subjects for validation, and 10,000 subjects for testing in this study.

\paragraph{BraTS2021}
To show the value of Seq2Seq, totipotency and imaging-differentiation representation in real practice, We also utilize brain MRI images of 1,251 patients from Brain Tumor Segmentation 2021 (BraTS2021)~\cite{baid2021rsna,bakas2017advancing,menze2014multimodal}. For each patient, four sequences are available, including T1, T1Gd, T2, and Flair. All the subjects are annotated with semantically meaningful tumor segmentation masks and a subset of 574 subjects are labeled with the MGMT promoter methylation status.
To mimic a clinical situation with complex incomplete data, we randomly silence (make it unavailable) one-to-three sequences for each subject to make an incomplete subset. We use 830, 93, and 328 patients for training, validation, and testing, respectively. All the images are normalized to [0, 1] after min-max clipping and then central-crop to a size of $128\times192\times192$.

\paragraph{In-house dataset}
We also collect an in-house dataset of 2,101 patients with breast cancer who are treated with neoadjuvant therapy (NAT) between 2000 to 2020 within a cancer institute. Two-to-four visits of MRI (pre-NAT, during NAT, and post-NAT) are retrieved for each patient. The acquired breast MRI protocol is varied due to the long time span of inclusion. Scanning before 2015 is done using 2D sequences in the coronal plane, but after 2015 shifted to 3D scanning in the transverse plane. Fat-saturation techniques are only employed for T1 and DCE in later years. Overall, eight sequences are present in the in-house dataset, including T1, fat-saturated T1 (FatSatT1), T2, DCE, fat-saturated DCE (FatSatDCE), wash-in, wash-out, and DWI. Among them, DCE, FatSatDCE, and DWI are 4D MRIs. DCE and FatSatDCE are acquired with 6 timepoints within 7.5 min after contrast injection. DWI consists of 4 different $b$ values of 0, 150, 800, and 1,500.
The MRIs are acquired with Philips Ingenia 3.0-T scanners. All the data are collected and utilized with the approval of the local ethics committee. We also collect pathological complete response (pCR) status for all the patients after NAT.
We randomly select 1,701 patients for training, 189 patients for validation, and 211 patients for testing.
All the images are resampled to $1\times1\times1$ mm and then cropped to keep the bilateral breast area with the size of $176\times176\times352$ for the coronal plane (without thoracic cavity) and with the size of $176\times352\times352$ for the transverse plane (with thoracic cavity).

\paragraph{Implementation details} \label{sec:implementation}
The models are developed with PyTorch and trained on the NVIDIA Quadro RTX A6000 GPU. For BraTS2021 and the in-house dataset, the Seq2Seq generator is trained using Adam optimizer with 1,000,000 steps, a batch size of 1, and a learning rate of $2\times10^{-4}$.
For easier optimization, we train the Conv2d model in 2.5D mode – three adjacent axial slices are treated as a three-channel 2D input. Segmentation by nnU-Net~\cite{isensee2021nnu} is trained with its default 3D full-resolution configuration. Classification by pre-trained 3D ResNet-18~\cite{chen2019med3d} is optimized with Adam, 50 epochs, a batch size of 2, and a learning rate of $10^{-3}$. For the toy dataset, we train the Seq2Seq generator with 10 epochs and optimize ResNet-18~\cite{he2016deep} with a batch size of 128.

\subsection{Sequence generation} \label{sec:seq_gen}
We compare our method with Pix2Pix~\cite{isola2017image}, MM-GAN~\cite{sharma2019missing}, ResViT~\cite{dalmaz2022resvit}, and other ablation study methods with the metrics of peak signal noise rate (PSNR), structural similarity index measure (SSIM), learned
perceptual image patch similarity (LPIPS)~\cite{zhang2018unreasonable}, Mega parameter numbers (MPa), and Giga multiply-accumulate operations per second (GMACs) on BraTS2021 and the in-house dataset, respectively.
For Pix2Pix~\cite{isola2017image}, we utilize a separate model for each pair of sequence translations, \textit{i.e.} 12 models for BraTS2021 (4 sequences) and 56 models for the in-house dataset (8 sequences).
We also compare different versions of our Seq2Seq model, equipped with AdaIN~\cite{huang2017arbitrary}, dynamic filters~\cite{shen2018neural}, and our proposed HyperConv layer. These three models are named Seq2Seq-A, Seq2Seq-B, and Seq2Seq-C. To simulate clinical settings with incomplete sequences, we utilize the BraTS2021 with random silence to train the models.
Table~\ref{tab:results} shows the sequence generation performance of comparison methods. Our proposed Seq2Seq-C achieves the best results of PSNR, SSIM, and LPIPS, and also has the smallest MPa and temperate GMACs. The number of models for Pix2Pix~\cite{isola2017image} is increasing sharply with the rise of the number of sequences, and cannot learn correlation information from a limited dataset. Seq2Seq-A and B have poor results of generation and also have high MPa and GMACs due to the drawback of AdaIN~\cite{huang2017arbitrary} and dynamic filters~\cite{shen2018neural}.
More specific results and visualization of sequence generation can be found in the supplementary materials.

\begin{table}
	\centering
	\caption{The quantitative results and profiles of comparison models for sequence generation on BraTS2021 and in-house dataset. The best result is in bold and the second best one is underlined.}
	\label{tab:results}
	\setlength{\tabcolsep}{3pt}
	\begin{tabular}{lccccccccccc}
		\toprule
		\multirow{2}*{Method}
            & \multicolumn{5}{c}{BraTS2021} && \multicolumn{5}{c}{In-house} \\
            \cline{2-6}\cline{8-12}
            & PSNR$\uparrow$ & SSIM$\uparrow$ & LPIPS$\downarrow$ & MPa$\downarrow$ & GMACs$\downarrow$ && PSNR$\uparrow$ & SSIM$\uparrow$ & LPIPS$\downarrow$ & MPa$\downarrow$ & GMACs$\downarrow$ \\
		\midrule
		Pix2Pix~\cite{isola2017image} & 26.4$\pm$3.2 & 0.849$\pm$0.067 & 13.8$\pm$4.5 & 136.5 & \underline{31.6} && 24.1$\pm$7.9 & 0.578$\pm$0.315 & 30.7$\pm$16.5 & 638.5 & \textbf{55.0} \\
            MM-GAN~\cite{sharma2019missing} & \underline{27.4$\pm$3.3} & \underline{0.863$\pm$0.068} & \underline{11.6$\pm$3.8} & \textbf{29.2} & \textbf{10.0} && - & - & - & - & - \\
            ResViT~\cite{dalmaz2022resvit} & 26.8$\pm$3.1 & 0.857$\pm$0.067 & 12.0$\pm$3.5 & 1172.7 & 486.7 && - & - & - & - & - \\
		Seq2Seq-A & 20.8$\pm$1.7 & 0.465$\pm$0.077 & 18.5$\pm$4.1 & 43.7 & 105.5 && 23.4$\pm$6.2 & 0.474$\pm$0.339 & 33.9$\pm$16.0 & \underline{43.7} & 1,063.6 \\
		Seq2Seq-B & 27.0$\pm$3.0 & 0.846$\pm$0.061 & 12.9$\pm$4.1 & 520.6 & 83.8 && \underline{27.3$\pm$4.8} & \underline{0.675$\pm$0.179} & \underline{24.2$\pm$12.5} & 520.6 & 841.1 \\
		Seq2Seq-C & \textbf{27.8$\pm$3.3} & \textbf{0.872$\pm$0.062} & \textbf{10.9$\pm$4.1} & \underline{35.8} & 83.4 && \textbf{29.6$\pm$5.8} & \textbf{0.796$\pm$0.143} & \textbf{17.6$\pm$12.9} & \textbf{35.8} & \underline{840.6} \\
		\bottomrule
	\end{tabular}
\end{table}

\subsection{Sequence contribution analysis} \label{sec:selection}
Table~\ref{tab:degree-centrality} shows the result of the imaging-differentiation and totipotent degree centrality for each sequence in brain imaging (BraTS2021) and breast imaging (in-house dataset). For brain imaging, $\mathcal{C}_{t}$ of T1Gd (-0.410) is slightly higher than the other three sequences. Similar $\mathcal{C}_{t}$ for these sequences indicate that none of them is sufficient to generate all the sequences perfectly. T1Gd and Flair achieve the top 2 $\mathcal{C}_{d}$ of 1.734 and 1.455, which present obvious imaging differences compared to T1 and T2.
For breast imaging, DCE and FatSatDCE have the best two $\mathcal{C}_{t}$ of 2.013 and 1.199, which are much higher than other sequences. It makes sense because T1/FatSatT1, wash-in, and wash-out can be calculated from DCE/FatSatDCE. There is no doubt that DCE, which is the main sequence for breast cancer evaluation, gets the highest $\mathcal{C}_{d}$ of 2.741 due to the most plentiful imaging-differentiation details. As the fat-saturation technique suppresses the fat signal, $\mathcal{C}_{d}$ of FatSatDCE is lower than that of DCE but is still higher than that of FatSatT1. T2 and DWI also achieve high ranks of $\mathcal{C}_{d}$ and indeed are used as a complement to DCE in the standard clinical protocol for cysts, mucinous carcinoma, necrotic cancer, and metaplastic carcinoma~\cite{mann2019breast}.

\begin{table}
	\centering
	\caption{The degree centrality and corresponding rank of imaging differentiation and totipotency for each sequence in BraTS2021 and in-house dataset.}
	\label{tab:degree-centrality}
	\setlength{\tabcolsep}{3pt}
	\begin{tabular}{cccccc}
		\toprule
		Dataset & Sequence & $\mathcal{C}_{t}$ & ${\rm Rank}_t$ & $\mathcal{C}_{d}$ & ${\rm Rank}_d$ \\
		\midrule
		\multirow{4}*{BraTS2021}
		    & T1 & -0.881 & 4 & -0.955 & 4 \\
		    & T1Gd & -0.410 & 1 & 1.734 & 1 \\
		    & T2 & -0.518 & 2 & 0.123 & 3 \\
		    & Flair & -0.548 & 3 & 1.455 & 2 \\
		\midrule
		\multirow{8}*{In-house}
		    & T1 & -2.722 & 8 & 0.968 & 4 \\
		    & FatSatT1 & -0.261 & 3 & 0.443 & 6 \\
		    & T2 & -0.430 & 4 & 1.737 & 2 \\
		    & DCE & 2.013 & 1 & 2.741 & 1 \\
		    & FatSatDCE & 1.199 & 2 & 0.656 & 5 \\
		    & Wash-in & -1.266 & 5 & -0.370 & 7\\
		    & Wash-out & -1.334 & 6 & -0.596 & 8 \\
		    & DWI & -1.670 & 7 & 1.516 & 3 \\
		\bottomrule
	\end{tabular}
\end{table}

To prove that sequences with higher $\mathcal{C}_{d}$ may make the ones ranked lower unnecessary in medical image analysis, we compare the tumor segmentation performance on BraTS2021 by training nnU-Net~\cite{isensee2021nnu} using different sequence combinations.
Complete sequences are used in this experiment and the split of the dataset is the same as that in Section~\ref{sec:seq_gen}.
Two metrics, DSC and ASSD, are utilized to evaluate the enhancing tumor (ET), the tumor core (TC), and the whole tumor (WT).
As shown in Table~\ref{tab:segmentation}, the combination of four sequences only achieves the best DSC of 0.923 and ASSD of 1.242 on WT but gets slightly worse performance on ET and TC.
Although more sequences bring more information, more interference is apparently also introduced.
A one-tailed paired sample T-test compares the complete sequences and other combinations.
The segmentation performance using only T1Gd and Flair is no worse than that of using complete sequences statistically ($p>0.05$).
The results are in good agreement with clinical experience~\cite{baid2021rsna}, that T1Gd describes ET and necrotic core of the tumor (NCR), and Flair defines the peritumoral edematous and infiltrated tissue (ED).

\begin{table}
	\centering
	\caption{Brain tumor segmentation results of nnU-Net on BraTS2021 by training with different MRI sequence combinations. The best result is in bold and the second best one is underlined. The one-sided paired samples T-test compares the results between complete sequences (last row) and other combinations. {*} marked on the results indicates a statistically significant difference of $p<0.05$.}
	\label{tab:segmentation}
	\setlength{\tabcolsep}{3pt}
	\begin{tabular}{ccccccccccc}
		\toprule
		\multirow{2}*{T1} & \multirow{2}*{T1Gd} & \multirow{2}*{T2} & \multirow{2}*{Flair}
            & \multicolumn{3}{c}{DSC$\uparrow$} && \multicolumn{3}{c}{ASSD$\downarrow$} \\
            \cline{5-7}\cline{9-11}
            &&&& ET & TC & WT && ET & TC & WT \\
            \midrule
            &\checkmark&&& \textbf{0.848$\pm$0.201} & {*}0.818$\pm$0.290 & {*}0.802$\pm$0.169 && \textbf{1.499$\pm$5.279} & {*}3.285$\pm$9.163 & {*}3.156$\pm$5.462 \\
            &&&\checkmark& {*}0.518$\pm$0.256 & {*}0.711$\pm$0.260 & {*}0.910$\pm$0.118 && {*}4.395$\pm$8.754 & {*}4.428$\pm$8.403 & 1.612$\pm$4.287 \\
            \checkmark&&\checkmark&& {*}0.563$\pm$0.253 & {*}0.705$\pm$0.270 & {*}0.884$\pm$0.112 && {*}3.700$\pm$7.082 & {*}3.963$\pm$6.974 & {*}1.837$\pm$3.954  \\
            &\checkmark&&\checkmark& \underline{0.843$\pm$0.213} & \textbf{0.844$\pm$0.248} & {*}\underline{0.917$\pm$0.102} && 1.925$\pm$7.376 & \textbf{2.341$\pm$6.919} & \underline{1.330$\pm$3.862} \\
            \checkmark&\checkmark&\checkmark&\checkmark& 0.837$\pm$0.210 & \underline{0.838$\pm$0.254} & \textbf{0.923$\pm$0.088} && \underline{1.785$\pm$5.957} & \underline{2.405$\pm$6.103} & \textbf{1.242$\pm$3.152} \\
		\bottomrule
	\end{tabular}
\end{table}

\subsection{Application of imaging-differentiation map}
\begin{figure}[!htbp]
    \centering
    \includegraphics[width=0.5\linewidth]{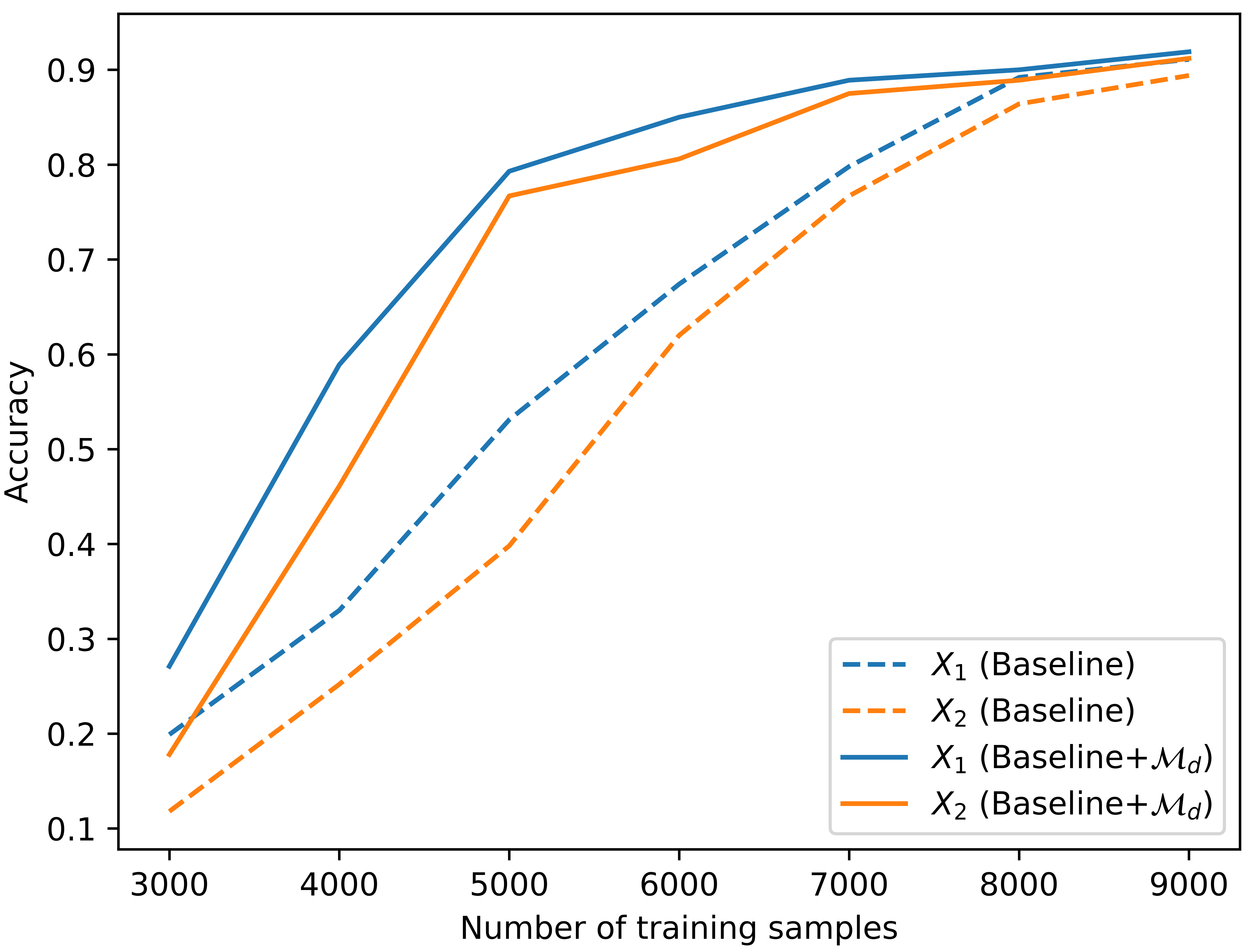}
    \caption{The curves between the 26-classification accuracy and the number of training samples. Blue and orange curves refer to the results of $X_1$ and $X_2$, respectively. The dotted lines present the results of Baseline and the solid line is the results of Baseline+$\mathcal{M}_d$.} \label{fig:prototype_curve}
\end{figure}

\paragraph{Alphabet recognition}
As shown in Fig.~\ref{fig:prototype}, the imaging-differentiation map focuses on the different alphabets between $X_1$ and $X_2$.
We further evaluate the role of the imaging-differentiation map on the toy dataset. We propose two models including (1) Baseline: ResNet-18~\cite{he2016deep} inputting with combination of $X_1$ and $X_2$; and (2) Baseline+$\mathcal{M}_d$: ResNet-18~\cite{he2016deep} inputting with combination of $X_1$, $X_2$, $\mathcal{M}_d(X_1)$, and $\mathcal{M}_d(X_2)$.
We train the model with different numbers of samples ranging from 3,000 to 9,000 and plot the curve between accuracy and sample number.
Fig.~\ref{fig:prototype_curve} illustrates that training with the imaging-differentiation map can improve the classification accuracy, and the improvement reaches the peak of 0.262 for $X_1$ and 0.369 for $X_2$ when the number of the training samples is 5000.

\begin{figure}[!htbp]
    \begin{minipage}{0.1\linewidth}
        \centerline{Case 1}
    \end{minipage}
    \begin{minipage}{0.1\linewidth}
        \centerline{\includegraphics[width=\textwidth]{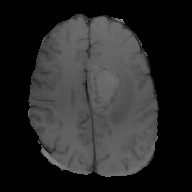}}
    \end{minipage}
    \begin{minipage}{0.1\linewidth}
        \centerline{\includegraphics[width=\textwidth]{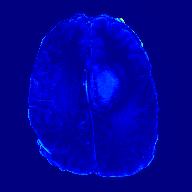}}
    \end{minipage}
    \begin{minipage}{0.1\linewidth}
        \centerline{\includegraphics[width=\textwidth]{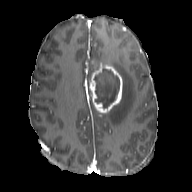}}
    \end{minipage}
    \begin{minipage}{0.1\linewidth}
        \centerline{\includegraphics[width=\textwidth]{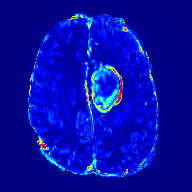}}
    \end{minipage}
    \begin{minipage}{0.1\linewidth}
        \centerline{\includegraphics[width=\textwidth]{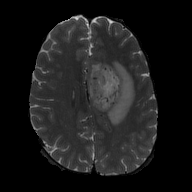}}
    \end{minipage}
    \begin{minipage}{0.1\linewidth}
        \centerline{\includegraphics[width=\textwidth]{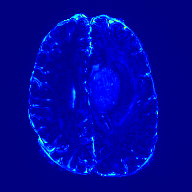}}
    \end{minipage}
    \begin{minipage}{0.1\linewidth}
        \centerline{\includegraphics[width=\textwidth]{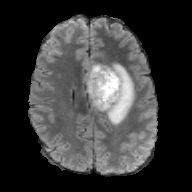}}
    \end{minipage}
    \begin{minipage}{0.1\linewidth}
        \centerline{\includegraphics[width=\textwidth]{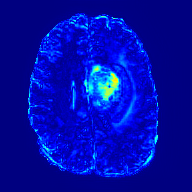}}
    \end{minipage}
    \begin{minipage}{0.018\linewidth}
        \includegraphics[width=\textwidth]{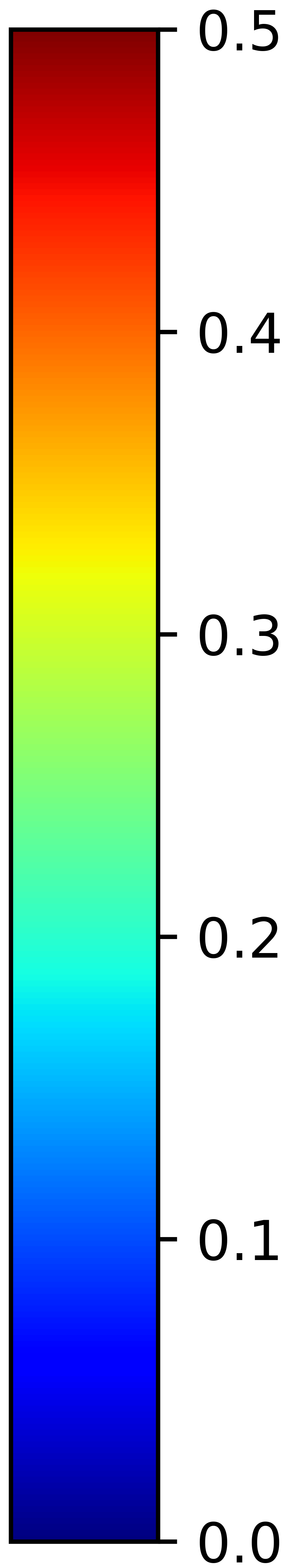}
    \end{minipage}
    
    
    \begin{minipage}{0.1\linewidth}
        \centerline{Case 2}
    \end{minipage}
    \begin{minipage}{0.1\linewidth}
        \centerline{\includegraphics[width=\textwidth]{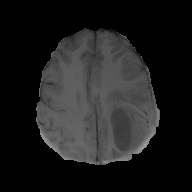}}
    \end{minipage}
    \begin{minipage}{0.1\linewidth}
        \centerline{\includegraphics[width=\textwidth]{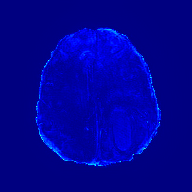}}
    \end{minipage}
    \begin{minipage}{0.1\linewidth}
        \centerline{\includegraphics[width=\textwidth]{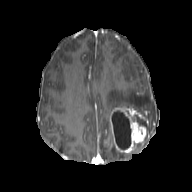}}
    \end{minipage}
    \begin{minipage}{0.1\linewidth}
        \centerline{\includegraphics[width=\textwidth]{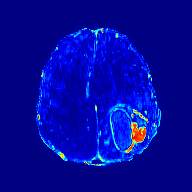}}
    \end{minipage}
    \begin{minipage}{0.1\linewidth}
        \centerline{\includegraphics[width=\textwidth]{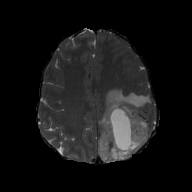}}
    \end{minipage}
    \begin{minipage}{0.1\linewidth}
        \centerline{\includegraphics[width=\textwidth]{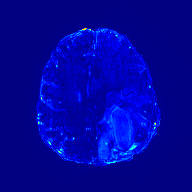}}
    \end{minipage}
    \begin{minipage}{0.1\linewidth}
        \centerline{\includegraphics[width=\textwidth]{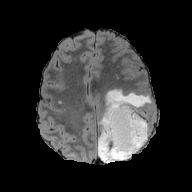}}
    \end{minipage}
    \begin{minipage}{0.1\linewidth}
        \centerline{\includegraphics[width=\textwidth]{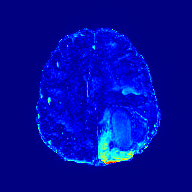}}
    \end{minipage}
    \begin{minipage}{0.018\linewidth}
        \includegraphics[width=\textwidth]{figs/colorbar_5.png}
    \end{minipage}
    
    
    \begin{minipage}{0.1\linewidth}
        \centerline{Case 3}
    \end{minipage}
    \begin{minipage}{0.1\linewidth}
        \centerline{\includegraphics[width=\textwidth]{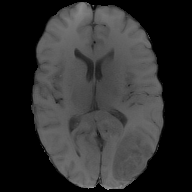}}
    \end{minipage}
    \begin{minipage}{0.1\linewidth}
        \centerline{\includegraphics[width=\textwidth]{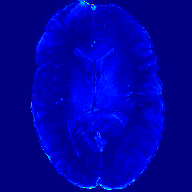}}
    \end{minipage}
    \begin{minipage}{0.1\linewidth}
        \centerline{\includegraphics[width=\textwidth]{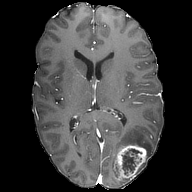}}
    \end{minipage}
    \begin{minipage}{0.1\linewidth}
        \centerline{\includegraphics[width=\textwidth]{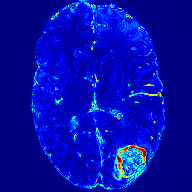}}
    \end{minipage}
    \begin{minipage}{0.1\linewidth}
        \centerline{\includegraphics[width=\textwidth]{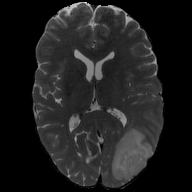}}
    \end{minipage}
    \begin{minipage}{0.1\linewidth}
        \centerline{\includegraphics[width=\textwidth]{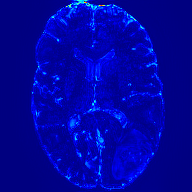}}
    \end{minipage}
    \begin{minipage}{0.1\linewidth}
        \centerline{\includegraphics[width=\textwidth]{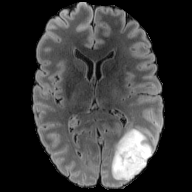}}
    \end{minipage}
    \begin{minipage}{0.1\linewidth}
        \centerline{\includegraphics[width=\textwidth]{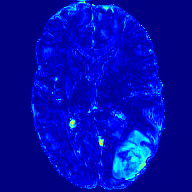}}
    \end{minipage}
    \begin{minipage}{0.018\linewidth}
        \includegraphics[width=\textwidth]{figs/colorbar_5.png}
    \end{minipage}
    
    
    \begin{minipage}{0.1\linewidth}
        \centerline{Case 4}
    \end{minipage}
    \begin{minipage}{0.1\linewidth}
        \centerline{\includegraphics[width=\textwidth]{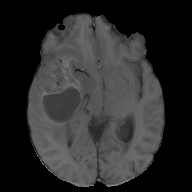}}
    \end{minipage}
    \begin{minipage}{0.1\linewidth}
        \centerline{\includegraphics[width=\textwidth]{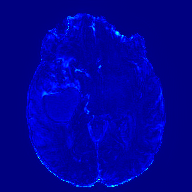}}
    \end{minipage}
    \begin{minipage}{0.1\linewidth}
        \centerline{\includegraphics[width=\textwidth]{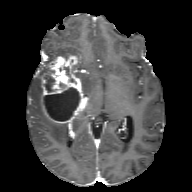}}
    \end{minipage}
    \begin{minipage}{0.1\linewidth}
        \centerline{\includegraphics[width=\textwidth]{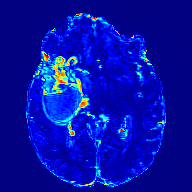}}
    \end{minipage}
    \begin{minipage}{0.1\linewidth}
        \centerline{\includegraphics[width=\textwidth]{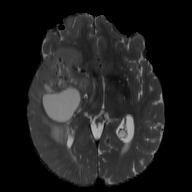}}
    \end{minipage}
    \begin{minipage}{0.1\linewidth}
        \centerline{\includegraphics[width=\textwidth]{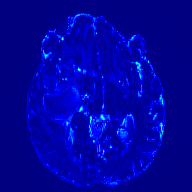}}
    \end{minipage}
    \begin{minipage}{0.1\linewidth}
        \centerline{\includegraphics[width=\textwidth]{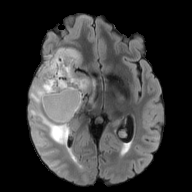}}
    \end{minipage}
    \begin{minipage}{0.1\linewidth}
        \centerline{\includegraphics[width=\textwidth]{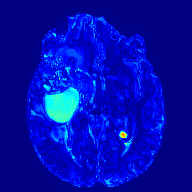}}
    \end{minipage}
    \begin{minipage}{0.018\linewidth}
        \includegraphics[width=\textwidth]{figs/colorbar_5.png}
    \end{minipage}
    
    \vspace{2pt}
    
    \begin{minipage}{0.1\linewidth}
        \centerline{}
    \end{minipage}
    \begin{minipage}{0.205\linewidth}
        \centerline{T1}
    \end{minipage}
    \begin{minipage}{0.205\linewidth}
        \centerline{T1Gd}
    \end{minipage}
    \begin{minipage}{0.205\linewidth}
        \centerline{T2}
    \end{minipage}
    \begin{minipage}{0.205\linewidth}
        \centerline{Flair}
    \end{minipage}
    \begin{minipage}{0.018\linewidth}
        \centerline{}
    \end{minipage}
	
	\caption{Examples of MRIs and corresponding heatmap of the imaging-differentiation map for different sequences in BraTS2021.} \label{fig:brats2021}
\end{figure}

\paragraph{MGMT promoter methylation prediction} 
Methylation of the MGMT promoter in newly diagnosed glioblastoma contributes to prognosis and has been identified as a predictor of response to chemotherapy. As shown in Fig.~\ref{fig:brats2021}, $\mathcal{M}_d$ highlights different regions of tumor in T1Gd and Flair, which is considered to be a guide for prediction. Here, we compare the prediction performance of MGMT promoter methylation status by training ResNet-18~\cite{chen2019med3d} with different combinations of input images, including (1) T1+T2; (2) T1+T2+$\mathcal{M}_d$; (3) T1Gd+Flair; (4) T1Gd+Flair+$\mathcal{M}_d$; (5) All; and (6) All+$\mathcal{M}_d$. In this section, the subset of BraTS2021 of 574 subjects with MGMT promoter methylation status is selected in the experiment. A 5$\times$4 two-level nested cross-validation is employed based on subjects to select the best model and reduce biased evaluation.
As shown in Table~\ref{tab:MGMT}, there are increments of AUC and accuracy by employing $\mathcal{M}_d$ into the model, and the improvement of AUC reaches the peak of 0.023 when using all the sequences.

\begin{table}
    \centering
    \caption{The results of predicting MGMT promoter methylation status. The best result is in bold and the second best one is underlined.}
    \label{tab:MGMT}
    \setlength{\tabcolsep}{3pt}
    \begin{tabular}{lcccccc}
        \toprule
        Method & AUC$\uparrow$ & Accuracy$\uparrow$ & Sensitivity$\uparrow$ & Specificity$\uparrow$ & PPV$\uparrow$ & NPV$\uparrow$ \\
        \midrule
        T1+T2 & 0.563$\pm$0.040 & 0.533$\pm$0.021 & 0.594$\pm$0.062 & 0.476$\pm$0.092 & 0.555$\pm$0.064 & 0.515$\pm$0.052 \\
        T1+T2+$\mathcal{M}_d$ & 0.575$\pm$0.038 & 0.549$\pm$0.033 & 0.533$\pm$0.170 & \textbf{0.574$\pm$0.166} & \textbf{0.587$\pm$0.075} & 0.534$\pm$0.048 \\
        T1Gd+Flair & 0.573$\pm$0.027 & 0.542$\pm$0.031 & \underline{0.673$\pm$0.148} & 0.430$\pm$0.167 & 0.568$\pm$0.080 & 0.559$\pm$0.091 \\
        T1Gd+Flair+$\mathcal{M}_d$ & \underline{0.585$\pm$0.027} & \underline{0.552$\pm$0.018} & 0.670$\pm$0.145 & 0.444$\pm$0.126 & 0.568$\pm$0.057 & \underline{0.576$\pm$0.114} \\
        All & 0.565$\pm$0.036 & 0.550$\pm$0.027 & 0.538$\pm$0.189 & \underline{0.554$\pm$0.174} & 0.576$\pm$0.060 & 0.529$\pm$0.047 \\
        All+$\mathcal{M}_d$ & \textbf{0.588$\pm$0.020} & \textbf{0.565$\pm$0.021} & \textbf{0.680$\pm$0.140} & 0.453$\pm$0.148 & \underline{0.578$\pm$0.068} & \textbf{0.576$\pm$0.081} \\
        \bottomrule
    \end{tabular}
\end{table}

\begin{figure}[!htbp]
    \begin{minipage}{0.18\linewidth}
        \centerline{\includegraphics[width=\textwidth]{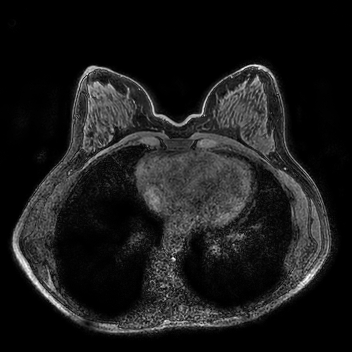}}
    \end{minipage}
    \begin{minipage}{0.18\linewidth}
        \centerline{\includegraphics[width=\textwidth]{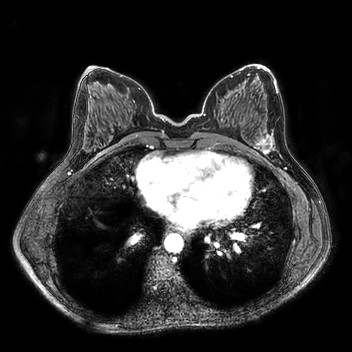}}
    \end{minipage}
    \begin{minipage}{0.18\linewidth}
        \centerline{\includegraphics[width=\textwidth]{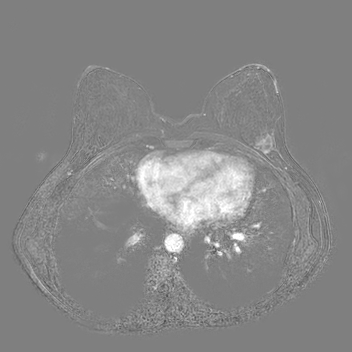}}
    \end{minipage}
    \begin{minipage}{0.18\linewidth}
        \centerline{\includegraphics[width=\textwidth]{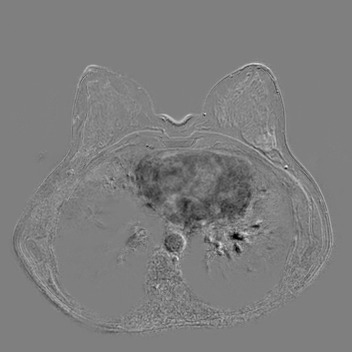}}
    \end{minipage}
    \begin{minipage}{0.18\linewidth}
        \centerline{\includegraphics[width=\textwidth]{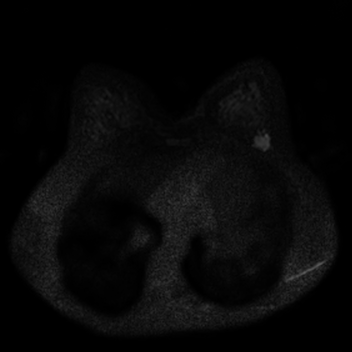}}
    \end{minipage}
    \begin{minipage}{0.03\linewidth}
        \centerline{}
    \end{minipage}
    
    
    \begin{minipage}{0.18\linewidth}
        \centerline{\includegraphics[width=\textwidth]{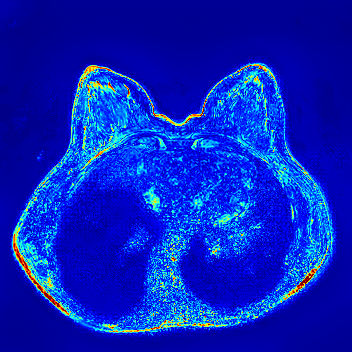}}
    \end{minipage}
    \begin{minipage}{0.18\linewidth}
        \centerline{\includegraphics[width=\textwidth]{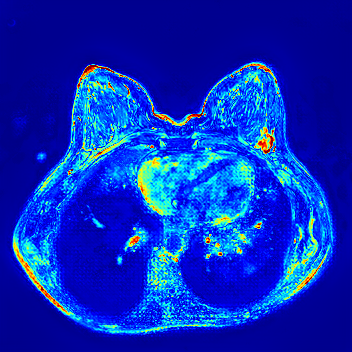}}
    \end{minipage}
    \begin{minipage}{0.18\linewidth}
        \centerline{\includegraphics[width=\textwidth]{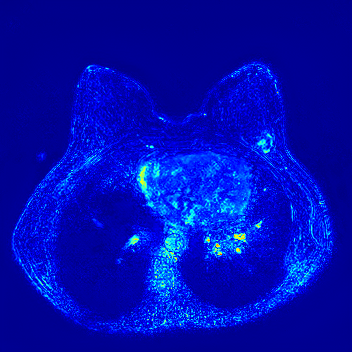}}
    \end{minipage}
    \begin{minipage}{0.18\linewidth}
        \centerline{\includegraphics[width=\textwidth]{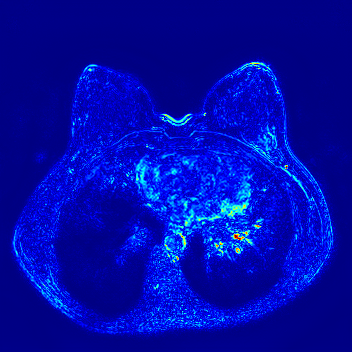}}
    \end{minipage}
    \begin{minipage}{0.18\linewidth}
        \centerline{\includegraphics[width=\textwidth]{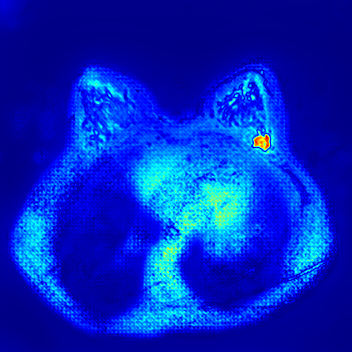}}
    \end{minipage}
    \begin{minipage}{0.03\linewidth}
        \vspace{15pt}
        \includegraphics[width=\textwidth]{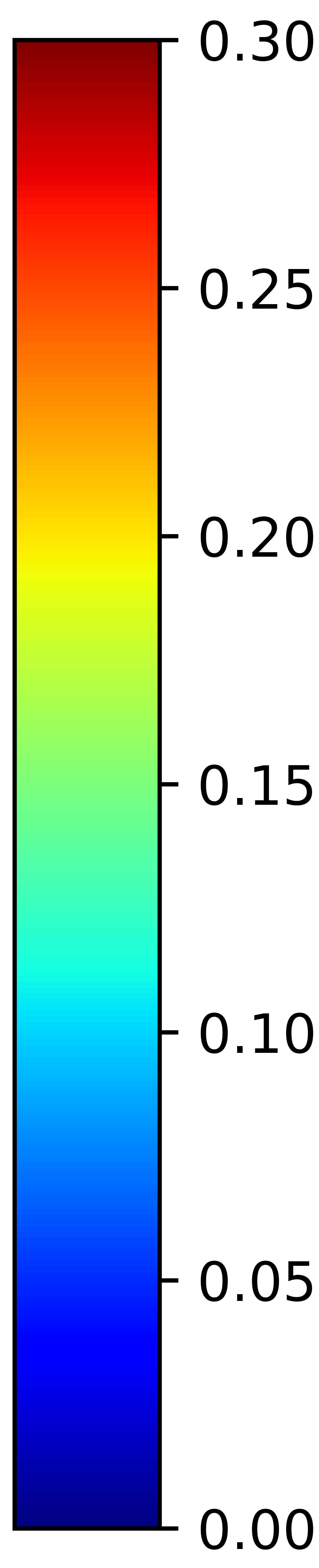}
    \end{minipage}
    
    
    \begin{minipage}{0.18\linewidth}
        \centerline{FatSatT1}
    \end{minipage}
    \begin{minipage}{0.18\linewidth}
        \centerline{FatSatDCE ($t$=1.5min)}
    \end{minipage}
    \begin{minipage}{0.18\linewidth}
        \centerline{Wash-in}
    \end{minipage}
    \begin{minipage}{0.18\linewidth}
        \centerline{Wash-out}
    \end{minipage}
    \begin{minipage}{0.18\linewidth}
        \centerline{DWI ($b$=800)}
    \end{minipage}
    \begin{minipage}{0.03\linewidth}
        \centerline{}
    \end{minipage}
	
	\caption{Examples of MRIs and corresponding heatmap of the imaging-differentiation map for different sequences in the in-house dataset.} \label{fig:breast}
\end{figure}

\begin{figure}[!htbp]
	\centering
	\includegraphics[width=0.5\linewidth]{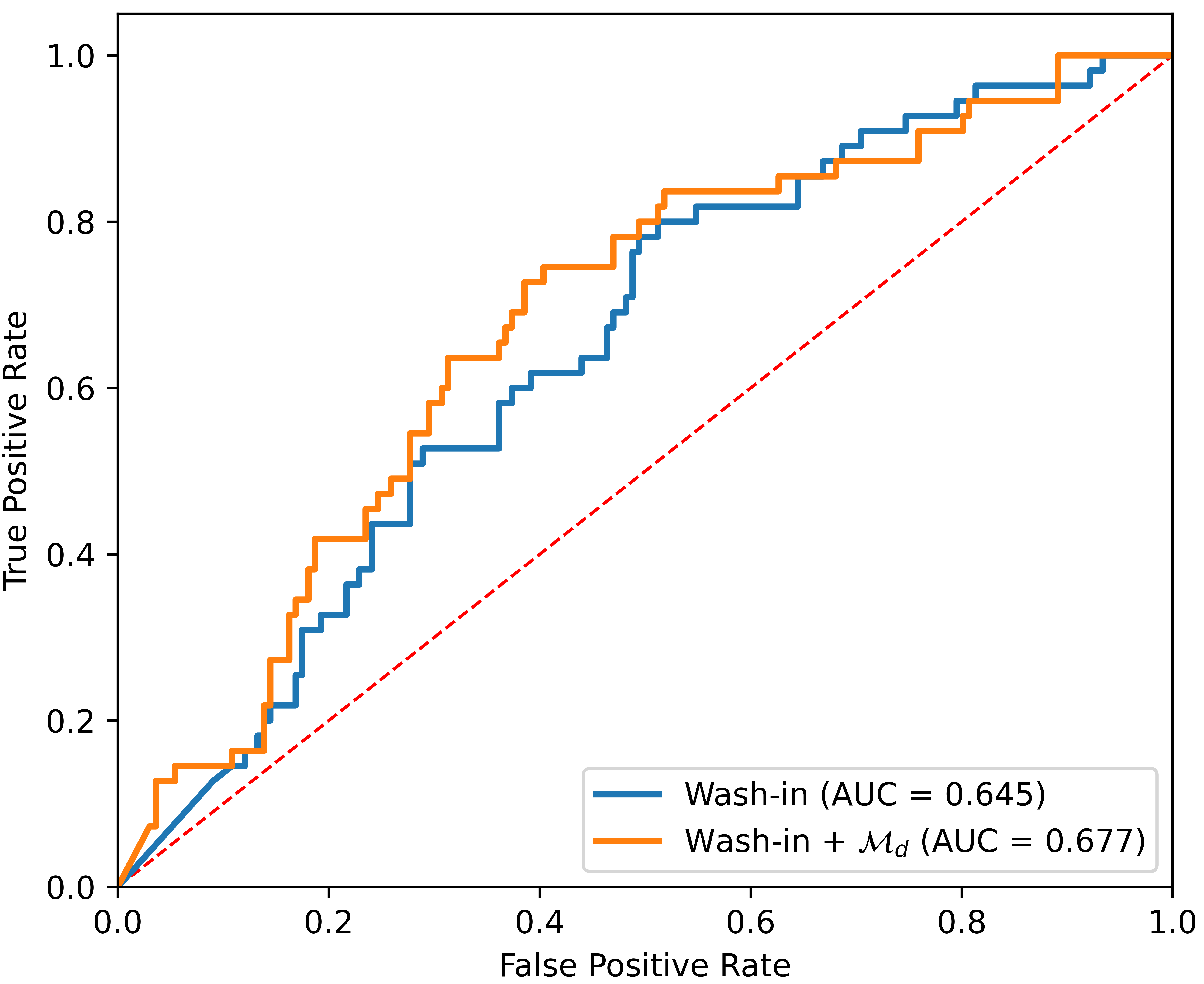}
	\caption{The ROC curve for pCR prediction with/without differentiation map on the in-house dataset.} \label{fig:roc}
\end{figure}

\paragraph{Breast cancer pCR early prediction}
Early prediction of the response to NAT for breast cancer can reduce ineffective treatments and adjust treatment plans in time.
Fig.~\ref{fig:breast} illustrates $\mathcal{M}_d$ for breast MRI, which is sensitive to the sequence-specific region.
Similarly, we train ResNet-18~\cite{chen2019med3d} to evaluate the improvement of pCR prediction before NAT by utilizing the corresponding $\mathcal{M}_d$. Due to incomplete sequences in the in-house dataset, we only utilize pre-NAT wash-in, which is not missing, to train the models in this section.
The experiment can be grouped as (1) wash-in and (2) wash-in + $\mathcal{M}_d$.
From Fig.~\ref{fig:roc}, before the treatment phase, the model trained with $\mathcal{M}_d$ can achieve better performance of pCR prediction with an increase of AUC of 0.032.

\section{Conclusion}
\label{sec:conclusion}
In this paper, we propose a one-for-all generation model to learn imaging-differentiation representation from arbitrary multi-sequence 3D/4D MRIs to generate any missing 3D/4D MRI sequences. Furthermore, for the first time, we qualitatively reveal the contribution of the sequence for diagnosis, and propose the sequence-specific regions from the imaging-differentiation maps to improve the downstream clinical applications such as MGMT promoter methylation status prediction and breast cancer pCR status prediction. Experimental results based on multiple datasets demonstrate our approach is efficient and reliable for multi-sequence MRI analysis.

\section*{Acknowledgments}
Luyi Han was funded by the Chinese Scholarship Council (CSC) scholarship.

\bibliographystyle{unsrt}  
\bibliography{references}

\begin{thebibliography}{10}

\bibitem{shukla2017advanced}
Gaurav Shukla, Gregory~S Alexander, Spyridon Bakas, Rahul Nikam, Kiran Talekar,
  Joshua~D Palmer, and Wenyin Shi.
\newblock Advanced magnetic resonance imaging in glioblastoma: a review.
\newblock {\em Chin Clin Oncol}, 6(4):40, 2017.

\bibitem{chen2013clinical}
Jeon-Hor Chen and Min-Ying Su.
\newblock Clinical application of magnetic resonance imaging in management of
  breast cancer patients receiving neoadjuvant chemotherapy.
\newblock {\em BioMed research international}, 2013, 2013.

\bibitem{feng2020knowledge}
Hongwei Feng, Jiaqi Cao, Hongyu Wang, Yilin Xie, Di~Yang, Jun Feng, and Baoying
  Chen.
\newblock A knowledge-driven feature learning and integration method for breast
  cancer diagnosis on multi-sequence mri.
\newblock {\em Magnetic resonance imaging}, 69:40--48, 2020.

\bibitem{grovik2020deep}
Endre Gr{\o}vik, Darvin Yi, Michael Iv, Elizabeth Tong, Daniel Rubin, and Greg
  Zaharchuk.
\newblock Deep learning enables automatic detection and segmentation of brain
  metastases on multisequence mri.
\newblock {\em Journal of Magnetic Resonance Imaging}, 51(1):175--182, 2020.

\bibitem{tang2020postoperative}
Fan Tang, Shujun Liang, Tao Zhong, Xia Huang, Xiaogang Deng, Yu~Zhang, and
  Linghong Zhou.
\newblock Postoperative glioma segmentation in ct image using deep feature
  fusion model guided by multi-sequence mris.
\newblock {\em European Radiology}, 30(2):823--832, 2020.

\bibitem{zhuang2022cardiac}
Xiahai Zhuang, Jiahang Xu, Xinzhe Luo, Chen Chen, Cheng Ouyang, Daniel
  Rueckert, Victor~M Campello, Karim Lekadir, Sulaiman Vesal, Nishant
  RaviKumar, et~al.
\newblock Cardiac segmentation on late gadolinium enhancement mri: a benchmark
  study from multi-sequence cardiac mr segmentation challenge.
\newblock {\em Medical Image Analysis}, 81:102528, 2022.

\bibitem{li2019diamondgan}
Hongwei Li, Johannes~C Paetzold, Anjany Sekuboyina, Florian Kofler, Jianguo
  Zhang, Jan~S Kirschke, Benedikt Wiestler, and Bjoern Menze.
\newblock Diamondgan: unified multi-modal generative adversarial networks for
  mri sequences synthesis.
\newblock In {\em Medical Image Computing and Computer Assisted
  Intervention--MICCAI 2019: 22nd International Conference, Shenzhen, China,
  October 13--17, 2019, Proceedings, Part IV 22}, pages 795--803. Springer,
  2019.

\bibitem{sharma2019missing}
Anmol Sharma and Ghassan Hamarneh.
\newblock Missing mri pulse sequence synthesis using multi-modal generative
  adversarial network.
\newblock {\em IEEE transactions on medical imaging}, 39(4):1170--1183, 2019.

\bibitem{zhou2020hi}
Tao Zhou, Huazhu Fu, Geng Chen, Jianbing Shen, and Ling Shao.
\newblock Hi-net: hybrid-fusion network for multi-modal mr image synthesis.
\newblock {\em IEEE transactions on medical imaging}, 39(9):2772--2781, 2020.

\bibitem{jung2021conditional}
Euijin Jung, Miguel Luna, and Sang~Hyun Park.
\newblock Conditional gan with an attention-based generator and a 3d
  discriminator for 3d medical image generation.
\newblock In {\em Medical Image Computing and Computer Assisted
  Intervention--MICCAI 2021: 24th International Conference, Strasbourg, France,
  September 27--October 1, 2021, Proceedings, Part VI 24}, pages 318--328.
  Springer, 2021.

\bibitem{uzunova2020memory}
Hristina Uzunova, Jan Ehrhardt, and Heinz Handels.
\newblock Memory-efficient gan-based domain translation of high resolution 3d
  medical images.
\newblock {\em Computerized Medical Imaging and Graphics}, 86:101801, 2020.

\bibitem{dalmaz2022resvit}
Onat Dalmaz, Mahmut Yurt, and Tolga {\c{C}}ukur.
\newblock Resvit: residual vision transformers for multimodal medical image
  synthesis.
\newblock {\em IEEE Transactions on Medical Imaging}, 41(10):2598--2614, 2022.

\bibitem{li2022virtual}
Wen Li, Haonan Xiao, Tian Li, Ge~Ren, Saikit Lam, Xinzhi Teng, Chenyang Liu,
  Jiang Zhang, Francis Kar-ho Lee, Kwok-hung Au, et~al.
\newblock Virtual contrast-enhanced magnetic resonance images synthesis for
  patients with nasopharyngeal carcinoma using multimodality-guided synergistic
  neural network.
\newblock {\em International Journal of Radiation Oncology* Biology* Physics},
  112(4):1033--1044, 2022.

\bibitem{han2018gan}
Changhee Han, Hideaki Hayashi, Leonardo Rundo, Ryosuke Araki, Wataru Shimoda,
  Shinichi Muramatsu, Yujiro Furukawa, Giancarlo Mauri, and Hideki Nakayama.
\newblock Gan-based synthetic brain mr image generation.
\newblock In {\em 2018 IEEE 15th international symposium on biomedical imaging
  (ISBI 2018)}, pages 734--738. IEEE, 2018.

\bibitem{campello2019combining}
V{\'\i}ctor~M Campello, Carlos Mart{\'\i}n-Isla, Cristian Izquierdo, Steffen~E
  Petersen, Miguel A~Gonz{\'a}lez Ballester, and Karim Lekadir.
\newblock Combining multi-sequence and synthetic images for improved
  segmentation of late gadolinium enhancement cardiac mri.
\newblock In {\em International Workshop on Statistical Atlases and
  Computational Models of the Heart}, pages 290--299. Springer, 2019.

\bibitem{qi2020multi}
Mengke Qi, Yongbao Li, Aiqian Wu, Qiyuan Jia, Bin Li, Wenzhao Sun, Zhenhui Dai,
  Xingyu Lu, Linghong Zhou, Xiaowu Deng, et~al.
\newblock Multi-sequence mr image-based synthetic ct generation using a
  generative adversarial network for head and neck mri-only radiotherapy.
\newblock {\em Medical physics}, 47(4):1880--1894, 2020.

\bibitem{huang2022evidence}
Ling Huang, Thierry Denoeux, Pierre Vera, and Su~Ruan.
\newblock Evidence fusion with contextual discounting for multi-modality
  medical image segmentation.
\newblock In {\em International Conference on Medical Image Computing and
  Computer-Assisted Intervention}, pages 401--411. Springer, 2022.

\bibitem{finck2022uncertainty}
Tom Finck, Hongwei Li, Sarah Schlaeger, Lioba Grundl, Nico Sollmann, Benjamin
  Bender, Eva B{\"u}rkle, Claus Zimmer, Jan Kirschke, Bj{\"o}rn Menze, et~al.
\newblock Uncertainty-aware and lesion-specific image synthesis in multiple
  sclerosis magnetic resonance imaging: A multicentric validation study.
\newblock {\em Frontiers in neuroscience}, 16, 2022.

\bibitem{huang2017arbitrary}
Xun Huang and Serge Belongie.
\newblock Arbitrary style transfer in real-time with adaptive instance
  normalization.
\newblock In {\em Proceedings of the IEEE international conference on computer
  vision}, pages 1501--1510, 2017.

\bibitem{shen2018neural}
Falong Shen, Shuicheng Yan, and Gang Zeng.
\newblock Neural style transfer via meta networks.
\newblock In {\em Proceedings of the IEEE Conference on Computer Vision and
  Pattern Recognition}, pages 8061--8069, 2018.

\bibitem{huang2018multimodal}
Xun Huang, Ming-Yu Liu, Serge Belongie, and Jan Kautz.
\newblock Multimodal unsupervised image-to-image translation.
\newblock In {\em Proceedings of the European conference on computer vision
  (ECCV)}, pages 172--189, 2018.

\bibitem{lee2018diverse}
Hsin-Ying Lee, Hung-Yu Tseng, Jia-Bin Huang, Maneesh Singh, and Ming-Hsuan
  Yang.
\newblock Diverse image-to-image translation via disentangled representations.
\newblock In {\em Proceedings of the European conference on computer vision
  (ECCV)}, pages 35--51, 2018.

\bibitem{choi2018stargan}
Yunjey Choi, Minje Choi, Munyoung Kim, Jung-Woo Ha, Sunghun Kim, and Jaegul
  Choo.
\newblock Stargan: Unified generative adversarial networks for multi-domain
  image-to-image translation.
\newblock In {\em Proceedings of the IEEE conference on computer vision and
  pattern recognition}, pages 8789--8797, 2018.

\bibitem{karras2020analyzing}
Tero Karras, Samuli Laine, Miika Aittala, Janne Hellsten, Jaakko Lehtinen, and
  Timo Aila.
\newblock Analyzing and improving the image quality of stylegan.
\newblock In {\em Proceedings of the IEEE/CVF conference on computer vision and
  pattern recognition}, pages 8110--8119, 2020.

\bibitem{jia2016dynamic}
Xu~Jia, Bert De~Brabandere, Tinne Tuytelaars, and Luc~V Gool.
\newblock Dynamic filter networks.
\newblock {\em Advances in neural information processing systems}, 29, 2016.

\bibitem{liu2019learning}
Xihui Liu, Guojun Yin, Jing Shao, Xiaogang Wang, et~al.
\newblock Learning to predict layout-to-image conditional convolutions for
  semantic image synthesis.
\newblock {\em Advances in Neural Information Processing Systems}, 32, 2019.

\bibitem{zhao2018dynamic}
Fang Zhao, Jian Zhao, Shuicheng Yan, and Jiashi Feng.
\newblock Dynamic conditional networks for few-shot learning.
\newblock In {\em Proceedings of the European conference on computer vision
  (ECCV)}, pages 19--35, 2018.

\bibitem{wu2018dynamic}
Jialin Wu, Dai Li, Yu~Yang, Chandrajit Bajaj, and Xiangyang Ji.
\newblock Dynamic filtering with large sampling field for convnets.
\newblock In {\em Proceedings of the European Conference on Computer Vision
  (ECCV)}, pages 185--200, 2018.

\bibitem{shi2015convolutional}
Xingjian Shi, Zhourong Chen, Hao Wang, Dit-Yan Yeung, Wai-Kin Wong, and
  Wang-chun Woo.
\newblock Convolutional lstm network: A machine learning approach for
  precipitation nowcasting.
\newblock {\em Advances in neural information processing systems}, 28, 2015.

\bibitem{zhu2017unpaired}
Jun-Yan Zhu, Taesung Park, Phillip Isola, and Alexei~A Efros.
\newblock Unpaired image-to-image translation using cycle-consistent
  adversarial networks.
\newblock In {\em Proceedings of the IEEE international conference on computer
  vision}, pages 2223--2232, 2017.

\bibitem{baid2021rsna}
Ujjwal Baid, Satyam Ghodasara, Suyash Mohan, Michel Bilello, Evan Calabrese,
  Errol Colak, Keyvan Farahani, Jayashree Kalpathy-Cramer, Felipe~C Kitamura,
  Sarthak Pati, et~al.
\newblock The rsna-asnr-miccai brats 2021 benchmark on brain tumor segmentation
  and radiogenomic classification.
\newblock {\em arXiv preprint arXiv:2107.02314}, 2021.

\bibitem{bakas2017advancing}
Spyridon Bakas, Hamed Akbari, Aristeidis Sotiras, Michel Bilello, Martin
  Rozycki, Justin~S Kirby, John~B Freymann, Keyvan Farahani, and Christos
  Davatzikos.
\newblock Advancing the cancer genome atlas glioma mri collections with expert
  segmentation labels and radiomic features.
\newblock {\em Scientific data}, 4(1):1--13, 2017.

\bibitem{menze2014multimodal}
Bjoern~H Menze, Andras Jakab, Stefan Bauer, Jayashree Kalpathy-Cramer, Keyvan
  Farahani, Justin Kirby, Yuliya Burren, Nicole Porz, Johannes Slotboom, Roland
  Wiest, et~al.
\newblock The multimodal brain tumor image segmentation benchmark (brats).
\newblock {\em IEEE transactions on medical imaging}, 34(10):1993--2024, 2014.

\bibitem{isensee2021nnu}
Fabian Isensee, Paul~F Jaeger, Simon~AA Kohl, Jens Petersen, and Klaus~H
  Maier-Hein.
\newblock nnu-net: a self-configuring method for deep learning-based biomedical
  image segmentation.
\newblock {\em Nature methods}, 18(2):203--211, 2021.

\bibitem{chen2019med3d}
Sihong Chen, Kai Ma, and Yefeng Zheng.
\newblock Med3d: Transfer learning for 3d medical image analysis.
\newblock {\em arXiv preprint arXiv:1904.00625}, 2019.

\bibitem{he2016deep}
Kaiming He, Xiangyu Zhang, Shaoqing Ren, and Jian Sun.
\newblock Deep residual learning for image recognition.
\newblock In {\em Proceedings of the IEEE conference on computer vision and
  pattern recognition}, pages 770--778, 2016.

\bibitem{isola2017image}
Phillip Isola, Jun-Yan Zhu, Tinghui Zhou, and Alexei~A Efros.
\newblock Image-to-image translation with conditional adversarial networks.
\newblock In {\em Proceedings of the IEEE conference on computer vision and
  pattern recognition}, pages 1125--1134, 2017.

\bibitem{zhang2018unreasonable}
Richard Zhang, Phillip Isola, Alexei~A Efros, Eli Shechtman, and Oliver Wang.
\newblock The unreasonable effectiveness of deep features as a perceptual
  metric.
\newblock In {\em Proceedings of the IEEE conference on computer vision and
  pattern recognition}, pages 586--595, 2018.

\bibitem{mann2019breast}
Ritse~M Mann, Nariya Cho, and Linda Moy.
\newblock Breast mri: state of the art.
\newblock {\em Radiology}, 292(3):520--536, 2019.

\end{thebibliography}

\end{document}